\documentclass[runningheads]{llncs}
\usepackage[T1]{fontenc}

\usepackage{algorithmic}
\usepackage{algorithm}
\usepackage{multirow}

\usepackage{graphicx}
\usepackage{url}
\usepackage{textcomp}

\usepackage{amsmath}
\usepackage{graphicx}

\begin{document}
\title{ENCLIP: Ensembling and Clustering-Based Contrastive Language-Image Pretraining for Fashion Multimodal Search with Limited Data and Low-Quality Images
}
\titlerunning{ENCLIP}
%
\author{Prithviraj Purushottam Naik\orcidID{0000-0002-8599-183X}\and
Rohit Agarwal
}
\authorrunning{P.P. Naik and R. Agarwal}
%
\institute{ Bizom (Mobisy Technologies Private Limited)\\
	\email{prithviraj.naik@mobisy.com}\\ 
	\and
	Bizom (Mobisy Technologies Private Limited) \\
	\email{rohit@mobisy.com}
}
\maketitle              
%

\begin{abstract}
Multimodal search has revolutionized the fashion industry, providing a seamless and intuitive way for users
to discover and explore fashion items. Based on their preferences, style, or specific attributes, users can search
for products by combining text and image information. Text-to-image searches enable users to find visually
similar items or describe products using natural language.
This paper presents an innovative approach called ENCLIP, for enhancing the performance of the
Contrastive Language-Image Pretraining (CLIP) model, specifically in Multimodal Search targeted towards
the domain of fashion intelligence. This method focuses on addressing the challenges posed by limited data
availability and low-quality images. This paper proposes an algorithm that involves training and ensembling multiple
instances of the CLIP model, and leveraging clustering techniques to group similar images together. The
experimental findings presented in this study provide evidence of the effectiveness of the methodology. This
approach unlocks the potential of CLIP in the domain of fashion intelligence, where data scarcity and image
quality issues are prevalent.
Overall, the ENCLIP method represents a valuable contribution to the field of fashion intelligence and
provides a practical solution for optimizing the CLIP model in scenarios with limited data and low-quality
images.\par

\end{abstract}
\section{Introduction}
The fashion industry is highly visual and dynamic, driven by constantly evolving trends and consumer preferences. Over 300 million people have jobs in the fashion industry, which generates global GDP of \$1.3 trillion \cite{1}. The rising prominence of e-commerce and online shopping platforms highlighted the need to develop fast and effective fashion search strategies. Traditional text-based search systems often fall short in capturing the rich visual aspects of fashion items, leading to suboptimal search results and a less satisfactory user experience \cite{2}.
To address these limitations, multimodal search \cite{3} has emerged as a powerful solution in the fashion domain. By leveraging both textual and visual information, multimodal search enables users to explore and discover fashion items in a more intuitive and comprehensive manner. This approach integrates the capabilities of natural language processing with computer vision techniques in order to establish a connection between textual descriptions and visual representations within the domain of fashion. In multimodal fashion search, users can express their preferences or query fashion items using natural language descriptions. These descriptions can encompass various aspects, such as color, gender, pattern, style, age, or even specific attributes like sleeve length or neckline. \par


This paper explores the utilization of multimodal search techniques in the context of fashion intelligence. The primary objective is to enhance the efficacy of the CLIP (Contrastive Language-Image Pretraining) \cite{4} model specifically for low-resolution and low-quality photos in the fashion domain. CLIP (Contrastive Language-Image Pretraining) model, a state-of-the-art deep learning model developed by OpenAI that excels in understanding the relationship between images and text.This paper proposes an approach called ENCLIP, which involves ensembling outputs of multiple CLIP fine-tuned models and leveraging clustering techniques to enhance the fine-tuning process. By harnessing the power of ensemble learning and exploiting the benefits of clustering, the challenges posed by limited data availability and low-quality images are addressed.\par

In the study, the dataset titled "Fashion Product Images (Small)" by Param Aggarwal \cite{5} available on Kaggle is utilized. This dataset comprises fashion images and their corresponding textual descriptions or keywords with some Indian fashion products. The dataset offers a diverse range of fashion items, including traditional attire and jewellery.
By leveraging this curated dataset that also includes Indian fashion products, this paper also aims to enhance the relevance and accuracy of the model with Indian fashion trends, styles, and cultural nuances. Following data collection, the fashion images are preprocessed to ensure uniform size and format, facilitating compatibility with CLIP. Simultaneously, the textual descriptions or keywords are prepared for input to CLIP's language model component.  The vector database was employed to facilitate the storing and retrieval of images, with input text queries expressed in the form of vector embeddings.\par

\section{Related work}
This section reviews previous related works on Multimodal Search in fashion domain.\par

In 2023 Min Wang et al.\cite{6} proposed a study that focuses
on training a visual question-answering (VQA) \cite{15} system for
apparel in fashion photoshoot images using a large-scale
multimodal dataset. By employing diverse templates and emphasizing challenging concepts, they achieved a VQA model
surpassing human expert-level accuracy, demonstrating the
effectiveness of visual language models for their dataset. The use of diverse templates and challenging concepts might require extensive preprocessing and template creation, increasing the complexity of implementation and potentially making the system less flexible.\par

In 2022 Karin Sevegnani et al.\cite{7} proposed a model, WhisperLite. This model utilizes contrastive learning techniques to effectively extract user intent from natural language text. By doing so, it enhances the quality of recommendations for fashion products.\par

In 2022 Patrick John Chia et al.\cite{8} proposed GradREC,
a novel recommendation system that introduces explicit directionality through natural language, enabling zero-shot,
language-based comparative recommendations without the
need for explicitly defined labels or behavioral data. Zero-shot learning capabilities, while impressive, may still struggle with rare or unseen attributes, leading to less accurate recommendations in certain scenarios.
\par

In 2022 Patrick John Chia et al.\cite{9} proposed FashionCLIP, a CLIP-like model for the fashion industry that showcases its capabilities for retrieval, classification, and grounding. FashionCLIP, being a CLIP-like model,require significant data and resources for training and inference. This model is used in this study to compare results with ENCLIP approach along with the Pre-trained CLIP Model.\par

In 2019 Ivona Tautkute et al. \cite{10} proposed a multimodal
search engine DeepStyle. The author demonstrates the effectiveness of this methodology on two distinct and demanding datasets consisting of fashion items and furniture. The DeepStyle engine surpasses baseline approaches by a significant margin of 18-21\% on the tested datasets. The effectiveness of the DeepStyle engine depends on the availability of high-quality visual and linguistic signals, which may not always be accessible.
\par

In 2019 Gil Sadeh et al. \cite{11} proposed a multimodal
visual-textual search refinement method for fashion garments,
allowing intuitive, interactive retrieval of similar items based on image and textual refinement properties. Their joint embedding training scheme effectively manipulates catalog images semantically using a new training objective function, Mini-Batch Match Retrieval, outperforming triplet loss, and they demonstrate the integration of an attribute extraction module to enhance search performance. The joint embedding training scheme and semantic manipulation of catalog images could be complex to implement and fine-tune, requiring extensive experimentation.\par

\section{Background}
This section discusses Multimodal Search and CLIP as used in the study.

\subsection{Multimodal Search}

Multimodal search refers to a search approach that employs several methodologies in order to obtain pertinent outcomes \cite{3}. It is intended to mimic the flexibility and agility with which the human mind generates, processes, and rejects irrelevant ideas. It is a type of search that allows users to search for information using multiple modalities, such as text, images, audio, and video. This can be done by providing a single query that includes text, images, or other media, or by providing multiple queries, each in a different modality. 

By combining different types of input, multimodal search engines can provide more accurate and relevant results than traditional search engines \cite{14}.
The different ways that multimodal search can be implemented include:

\begin{itemize}
  \item \textbf{Feature-based}: This approach uses features extracted from each modality to represent the query and the documents. The features are then combined to calculate the similarity between the query and the documents.

\item \textbf{Model-based}: This approach uses a model to learn the relationship between the different modalities. The model can then be employed to predict the relevance of documents to a query.

\item \textbf{Hybrid}: This approach combines the feature-based and model-based approaches.
\end{itemize}

\subsection{CLIP}

\begin{figure}[]
  \centering
\includegraphics[width=8cm, height=4cm]{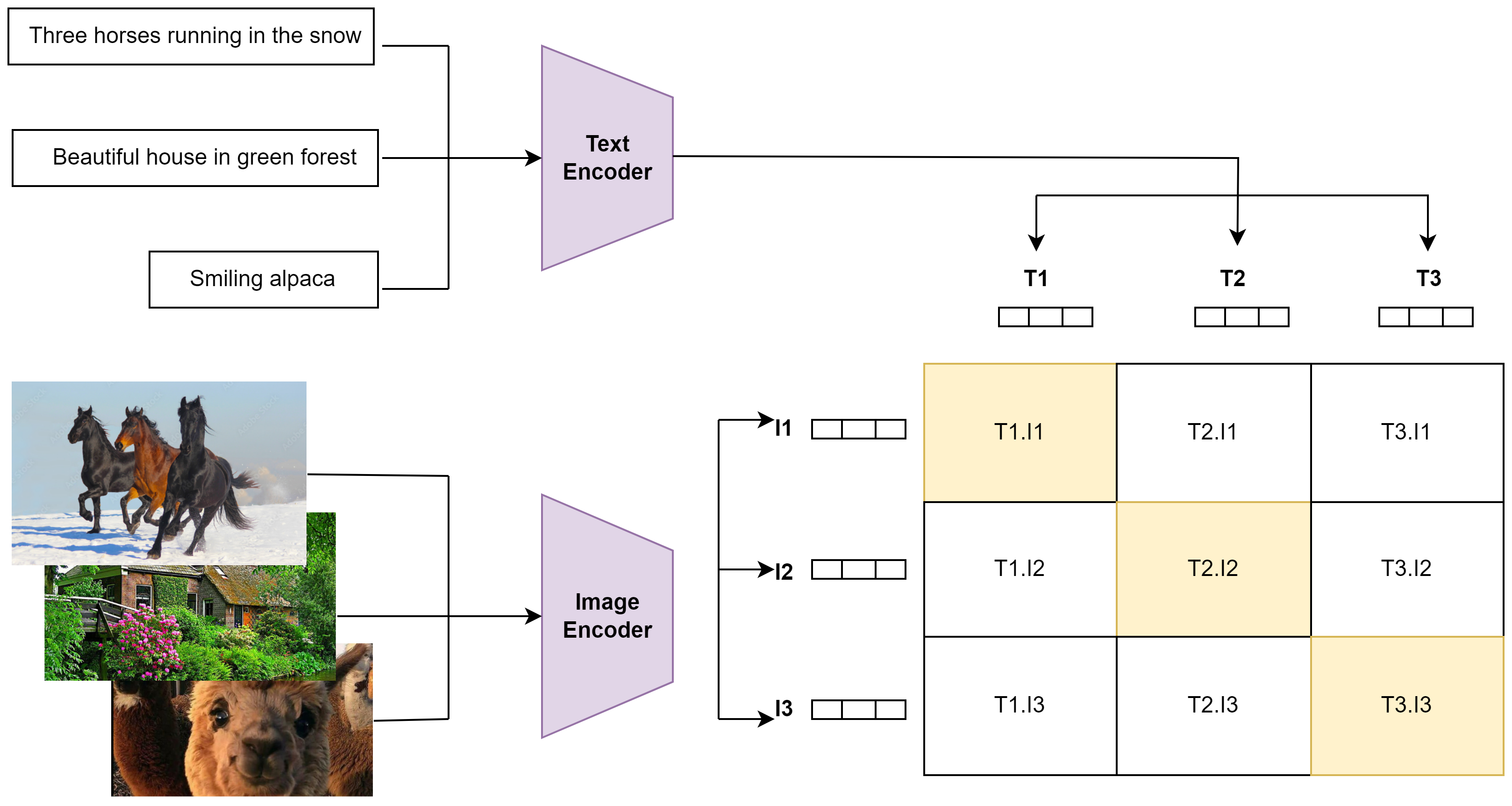}
  \caption{Contrastive pretraining with CLIP} 
  \label{1}
\end{figure}

The neural network model known as CLIP (Contrastive Language-Image Pre-training)\cite{4} was created by OpenAI. The model has been trained using a vast dataset consisting of pairings of images and corresponding texts. Fig. \ref{1} illustrates the concept of contrastive pretraining with CLIP.

CLIP is a powerful model with the potential to revolutionize the way images and text are interacted with. It is still under development, but it has already been used for a variety of interesting applications.

 The basic architecture of CLIP is as follows:

 \begin{itemize}
  \item The model is first pre-trained on a massive dataset of image-text pairs.
  \item The model then consists of two main components: a vision transformer \cite{16} and a language model \cite{17}.
  \item The vision transformer is responsible for encoding the images into a vector representation.
  \item The language model is responsible for encoding the text into a vector representation.
    \item The two vector representations are then compared to each other to determine the similarity between the image and the text.
\end{itemize}

The comparison of the two vector representations is done using a contrastive loss function \cite{18}. The contrastive loss function is designed to maximize the similarity between the image and text representations when they are actually related, and to minimize the similarity between the image and text representations when they are not related. The basic architecture of CLIP is relatively simple, but it is very effective at learning the semantic similarities between images and text.

\section{Methodology}
\subsection{Dataset}
This dataset, fashion-product-images-small \cite{5}, provided by HuggingFace, is used in this study. It consists of around 44072 60x80 pixel resolution images with details and descriptions about the image in respective columns. It has 7 master categories(Apparel, Accessories, Footwear, Personal Care, Free items, Home, Sporting Goods), 45 subcategories. 

\begin{figure}[]
\centerline{\includegraphics[width=8cm, height=2cm]{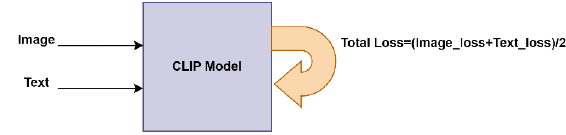}}
  \caption{Fine-tuning CLIP} 
  \label{3}
\end{figure}
\begin{figure}[]
\centerline{\includegraphics[width=8cm, height=2.5cm]{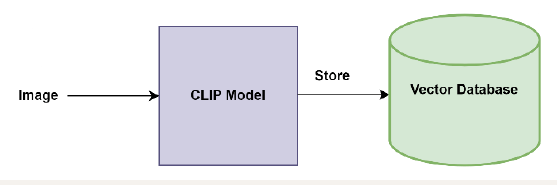}}
  \caption{Storing the image embedding and texual embedding in vector database} 
  \label{4}
\end{figure}

\begin{figure}[]
\centerline{\includegraphics[width=10cm, height=3.5cm]{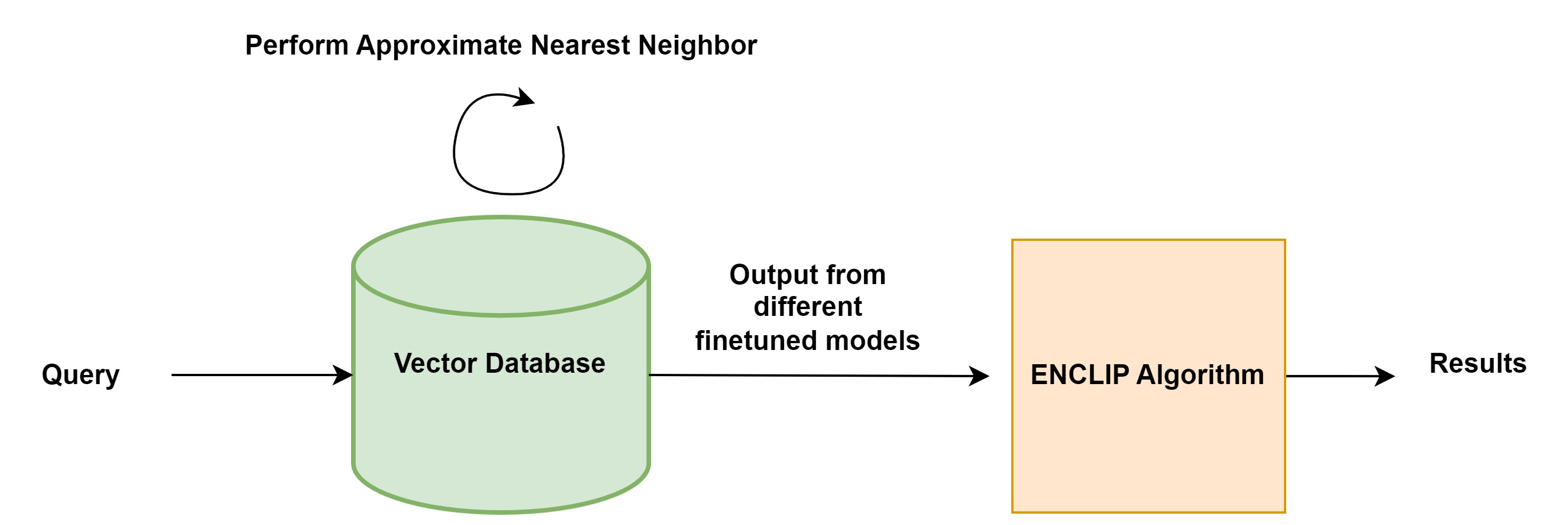}}
  \caption{Process of retrieving the results from vector database using ENCLIP algorithm. } 
  \label{5}
\end{figure}

\subsection{Preprocessing }
 
The dataset consists of around 44072 (image,text) pairs. It is divided into three sets of 80\% for training, 10\% for validation, and 10\% for testing. There are 35258 (image,text) pair in the training set,  4407 (image,text) pair in the validation set and 4407 (image,text) pair in the test set. The training model requires the images to be scaled from their original dimensions of 60x80 pixels to 224x224 pixels. All column values except for id and year are considered as captions for the corresponding image. Balanced Batch Sampling is performed as a sampling technique to ensure that each batch of data contains an equal representation of different classes. It is designed to address the issue of class imbalance, where some classes have significantly fewer samples than others, which can lead to biased model training.

\subsection{Model Architecture and Training}
Adamax is used as an optimization technique. Categorical cross-entropy was employed as a loss function for both image and text. Tokenization of the caption for an image was performed before giving it to the model for training. The Cosine Annealing learning rate scheduler, which gradually reduces the learning rate in a cosine annealing pattern, was used.

The cosine annealing pattern is designed to help the model converge to a better solution by allowing it to explore different areas of the loss landscape during training. It helps to avoid getting stuck in local minima and potentially find a more optimal solution. 

According to Equation \ref{2}, the total loss is calculated as the average of the image loss and text loss. \\
\begin{equation}
\text{total\,loss} = \frac{\text{image\,loss} + \text{text\,loss}}{2}
  \label{2}
\end{equation}

The model was trained across 10, 30, 50, 80, and 100 epochs, using a batch size of 128. It was executed on a Google Colab Pro platform equipped with an Nvidia Tesla T4 graphics card. During training, the model took an average of 280 seconds per epoch.

\subsection{Ensemble and Clustering Strategy}

\subsubsection{Rationale for Model Selection:}
Five CLIP models are trained, each initialized with the same architecture but trained for different epochs: 10, 30, 50, 80, and 100 (To demonstrate the working of the Algorithm \ref{6}, five fine-tuned models with incremental training are considered in this study). This approach captures various learning stages of the model, from initial learning to fine-tuning. It was observed that results in initial epoch training and final epoch training complemented each other to give the best results collectively.
Combining these models in an ensemble allows us to leverage the strengths of each. 

\subsubsection{Latent space of the Algorithm:}
Latent space is a lower-dimensional space that captures the essential features of the input data. In ENCLIP approach, each model encodes an image into a fixed-size vector (e.g., 512 dimensions), capturing the essential features and patterns in the data while reducing its dimensionality using t-SNE (t-distributed Stochastic Neighbor Embedding) to 2D for visualization purposes. This helps in understanding the distribution and structure of the data in a more interpretable form.

\subsubsection{Weighted Sum Approach:}
To combine the predictions of each model, a weighted sum approach is employed. The weights are determined based on the epoch number, with later epochs receiving higher weights:

\begin{equation}
weighted\_score=\sum_{n=0}^{z-1} 0.1\times(2^n)\times occurrence \
 \label{7}
\end{equation}

This weighting strategy ensures that more emphasis is placed on models that have undergone extensive training while still incorporating the insights from earlier training stages.

\subsubsection{Encoding Images:}
Each model is used to encode the images into a latent space, resulting in z sets of encoded images(z - Number of fine-tuned models; Five fine-tuned models (z=5) were considered in this study for demonstration of the Algorithm \ref{6}). These encoded representations capture the high-dimensional features learned by each model.

\subsubsection{Dimensionality Reduction:}
To visualize and analyze the encoded features, t-SNE (t-distributed Stochastic Neighbor Embedding) is applied, reducing the dimensionality to 2D while preserving the local structure of the data.

\subsubsection{Cluster Analysis and Ranking:}
K-Means clustering is performed on the t-SNE-transformed features to group similar images. The number of clusters (k) is set to values between 4 and 6, reflecting the diversity in the dataset categories.

\subsubsection{Cluster Label Assignment:}
Cluster labels are assigned to each image based on the K-Means results, and the frequency of each image appearing in different clusters is analyzed.

\subsubsection{Frequency and Weighted Sum Calculation:}
For each image, the frequency across different epochs is calculated, and a weighted sum based on predefined weights is computed. This helps in ranking the images according to their importance and relevance within each cluster.

\subsubsection{Final Selection:}
Images are ranked and selected based on their weighted sum and frequency, ensuring that the most representative and informative images are prioritized.

\begin{algorithm}
\caption{Ensemble-Cluster based Image Selection and Ranking Algorithm (ENCLIP Algorithm)}
\label{6}

\begin{algorithmic}[1]
\REQUIRE Image outputs from Fine-tuned models $M = {m_1, m_2, ..., m_z}$ , number of images to select $N$.
\ENSURE Selected and ranked $N$ images.
\STATE Perform ensembling by considering all fine-tuned models $M$ image outputs in the latent space.
\STATE Assign frequency values to the images based on its occurrence in each fine-tuned model's output.
\STATE Assign weighted scores for the images based on its occurrence in each fine-tuned model’s output as following:
\begin{equation}
weighted\_score=\sum_{n=0}^{z-1} 0.1\times(2^n)\times occurrence \
 \label{7}
\end{equation}
z = number of fine-tuned models used.
\STATE Apply K-means Clustering on all fine-tuned model's output in the latent space.
\STATE Initialize an empty list $selected\_heads$ to store head images.
\STATE Store all the  fine-tuned models image outputs in $selected\_heads$ and sort them in descending order based on $\text{frequency}$
\STATE Initialize an empty set $S$ to store the final selected images.
\STATE $i \gets 0$
\WHILE{$|S| < N$}
        \STATE Select the head image ($\text{selected\_heads}[i]$) and consider clusters with the head image as the head cluster.
        \STATE Select all the images in those head clusters and rank the images based on $\text{sort}(\text{frequency}, \textit{weighted\_score})$ and store them in $S$.
        \STATE $i \gets i\text{ + }1$
\ENDWHILE
\STATE Return the top $N$ images from $S$.
\end{algorithmic}
\end{algorithm}


\subsection{Evaluation Metrics}

In this section, the evaluation metrics used to measure the performance of the ENCLIP approach is discussed. \cite{12,13}.

\begin{itemize}
  \item \textbf{Precision(PREC)}: It can be defined as the percentage of relevant items out of those items selected by the query. 
  

\begin{equation}
PREC = \frac{N_{rs}}{N_{s}}
  \label{8}
\end{equation}

where ${N_{\text{rs}}}$ is the number of the recommended items that user prefer and ${N_{\text{s}}}$ is the number of the recommended items.\\

\item \textbf{Precision@k(PREC@k)}: It can be defined as the proportion of relevant recommended items in a recommendation list of size k for a query. 

\begin{equation}
PREC@k = \frac{N_{rs@k}}{k}
  \label{9}
\end{equation}

where ${N_{\text{rs@k}}}$ is the number of the recommended items that user prefer in a recommendation list of size k.\\

\item \textbf{Mean Average Precision(mAP)}: The mean average precision (MAP) is the average of multiple recommendation precision. The more precision of the recommended items are, the higher the mAP is, and the formula of mAP is as follows \cite{12}:


\begin{equation}
\text{AVG\_PREC@k}(q) = \frac{\sum_{k=1}^{n} \text{PREC@k} \times \text{rel}(k)}{\text{Number of relevant items }}
\label{10}
\end{equation}

\begin{equation}
\text{mAP} = \frac{1}{Q} \sum_{q=1}^{Q} \text{AVG\_PREC@k}(q)
\label{11}
\end{equation}

where Q is the number of recommendation, k is the rank, \text{rel}(k) represents the relativity function given rank k, \text{PREC@k}
represents the precision given rank k.

\end{itemize}

\section{Evaluation Results and Discussion}
Table.~\ref{12} and Table.~\ref{13} compare the retrieved results for the query "Give me polo neck t-shirt for men", with Table.~\ref{12} focusing on different fine-tuned CLIP models and Table.~\ref{13} comparing Pre-trained CLIP, FashionCLIP, and ENCLIP approaches. In Table.~\ref{13}, the following points are observed:
\begin{itemize}
\item In Pre-trained CLIP Model, the search output predominantly features full sleeved round neck t-shirts.
\item In FashionCLIP Model, the search output appears satisfactory but includes some full sleeved round neck t-shirts.
\item In ENCLIP Approach, the search output appears to be focused specifically on polo neck t-shirt for men.
\end{itemize}

\begin{table}[htbp]
\centering

\caption{Comparison between retrieved results from different CLIP fine-tuned models obtained by the query "Give me polo neck t-shirt for men"}
\begin{tabular}{|c|c|}
\hline
\textbf{Model} & \textbf{Retrieved Products} \\
\hline

Epoch 10 & \includegraphics[width=12cm, height=1cm]{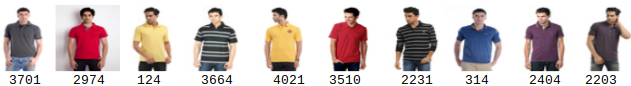} \\
\hline
Epoch 30 & \includegraphics[width=12cm, height=1cm]{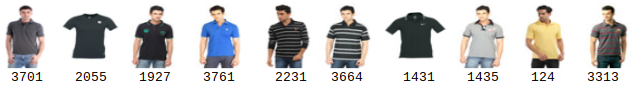} \\
\hline
Epoch 50 & \includegraphics[width=12cm, height=1cm]{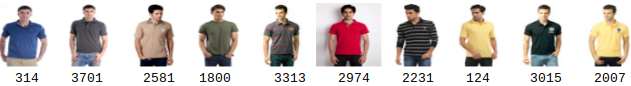} \\
\hline
Epoch 80 & \includegraphics[width=12cm, height=1cm]{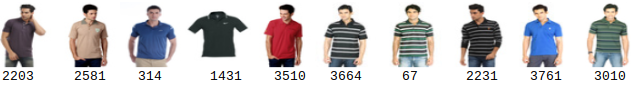} \\
\hline
Epoch 100 & \includegraphics[width=12cm, height=1cm]{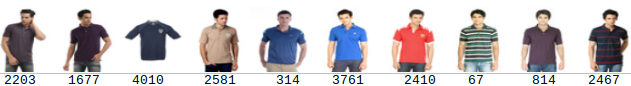} \\
\hline
\end{tabular}
\label{12}
\end{table}

\begin{table}[htbp]
\centering

\caption{Comparison between models for a sample of the qualitative results obtained by the query "Give me polo neck t-shirt for men"}
\begin{tabular}{|c|c|}
\hline
\textbf{Model} & \textbf{Retrieved Products} \\
\hline
Pre-trained CLIP & \includegraphics[width=11cm, height=1cm]{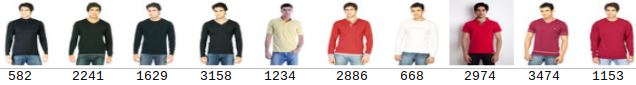} \\
\hline
FashionCLIP & \includegraphics[width=11cm, height=1cm]{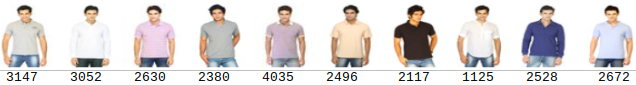} \\
\hline
ENCLIP Approach & \includegraphics[width=11cm, height=1cm]{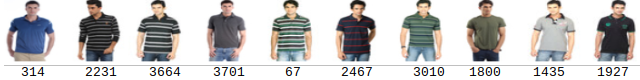} \\
\hline
\end{tabular}
\label{13}
\end{table}

\begin{table}[]
 \caption{Comparison of Mean Average Precision for Average Precision@10 for the FashionCLIP, Pre-trained CLIP and ENCLIP Approach.}
\begin{tabular}{|c|c|ccc|}
\hline
\multirow{2}{*}{\textbf{Category}}      & \multirow{2}{*}{\textbf{Sub Category}} & \multicolumn{3}{c|}{\textbf{mAP for AVG\_PREC@10}}                                                                    \\ \cline{3-5} 
                                        &                                        & \multicolumn{1}{c|}{\textbf{Pre-trained CLIP}} & \multicolumn{1}{c|}{\textbf{FashionCLIP}} & \textbf{ENCLIP Approach} \\ \hline
\multirow{2}{*}{Topwear and Bottomwear} & Topwear                                & \multicolumn{1}{c|}{0.552}                     & \multicolumn{1}{c|}{\textbf{0.797}}       & 0.651                    \\ \cline{2-5} 
                                        & Bottomwear                             & \multicolumn{1}{c|}{0.677}                     & \multicolumn{1}{c|}{0.657}                & \textbf{0.735}           \\ \hline
\multirow{2}{*}{Ethnic wear}            & Men                                    & \multicolumn{1}{c|}{0.062}                     & \multicolumn{1}{c|}{0.105}                & \textbf{0.185}          \\ \cline{2-5} 
                                        & Women                                  & \multicolumn{1}{c|}{0.773}                     & \multicolumn{1}{c|}{0.786}       & {\textbf{0.821}}                   \\ \hline
\multirow{2}{*}{Footwear}               & Men                                    & \multicolumn{1}{c|}{0.563}                     & \multicolumn{1}{c|}{0.811}                & \textbf{0.837}           \\ \cline{2-5} 
                                        & Women                                  & \multicolumn{1}{c|}{0.646}                     & \multicolumn{1}{c|}{0.683}                & \textbf{0.803}           \\ \hline
\multirow{2}{*}{Accessories}            & Men                                    & \multicolumn{1}{c|}{0.769}                     & \multicolumn{1}{c|}{0.934}                & \textbf{0.947}           \\ \cline{2-5} 
                                        & Women                                  & \multicolumn{1}{c|}{0.827}                     & \multicolumn{1}{c|}{\textbf{0.965}}       & 0.835                    \\ \hline
\multirow{2}{*}{Bags}                   & Men                                    & \multicolumn{1}{c|}{0.506}                     & \multicolumn{1}{c|}{0.598}                & \textbf{0.614}           \\ \cline{2-5} 
                                        & Women                                  & \multicolumn{1}{c|}{0.587}                     & \multicolumn{1}{c|}{0.611}                & \textbf{0.717}           \\ \hline
\end{tabular}
\label{14}
\end{table}

Table.~\ref{14} shows the comparison of Mean Average Precision for Average Precision@10 for the Pre-trained CLIP, FashionCLIP and ENCLIP Approach. In Table.~\ref{14} results, 10 queries for each category to calculate Mean Average Precision is considered. So, a total of 100 queries are considered for evaluation.

\begin{figure}[htbp]
\centerline{\includegraphics[width=8cm, height=5cm]{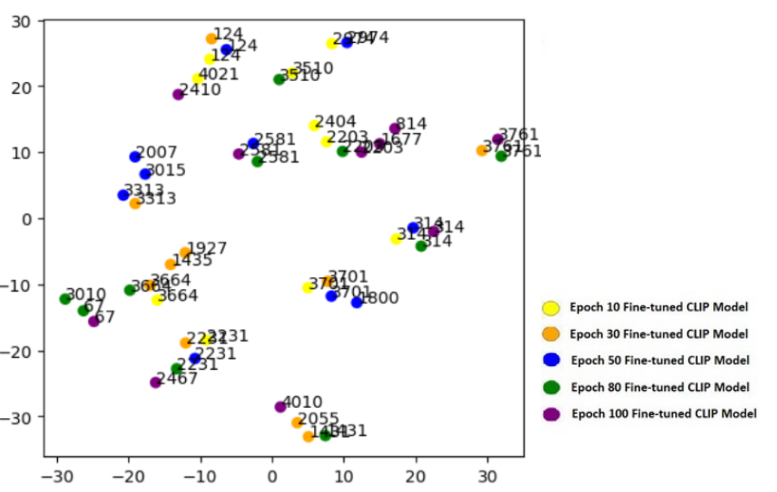}}
  \caption{t-SNE plot for different fine-tuned model's image outputs in same latent space  obtained by the query ”Give me polo neck t-shirt for men”} 
  \label{15}
\end{figure}

Fig.~\ref{15} represents the t-SNE plot for different fine-tuned model's image outputs in the same latent space obtained by the query ”Give me polo neck t-shirt for men”. Epoch 10, 30, 50, 80, and 100 fine-tuned CLIP models have been selected for this study.

\begin{figure}[htbp]
\centerline{\includegraphics[width=6cm, height=5cm]{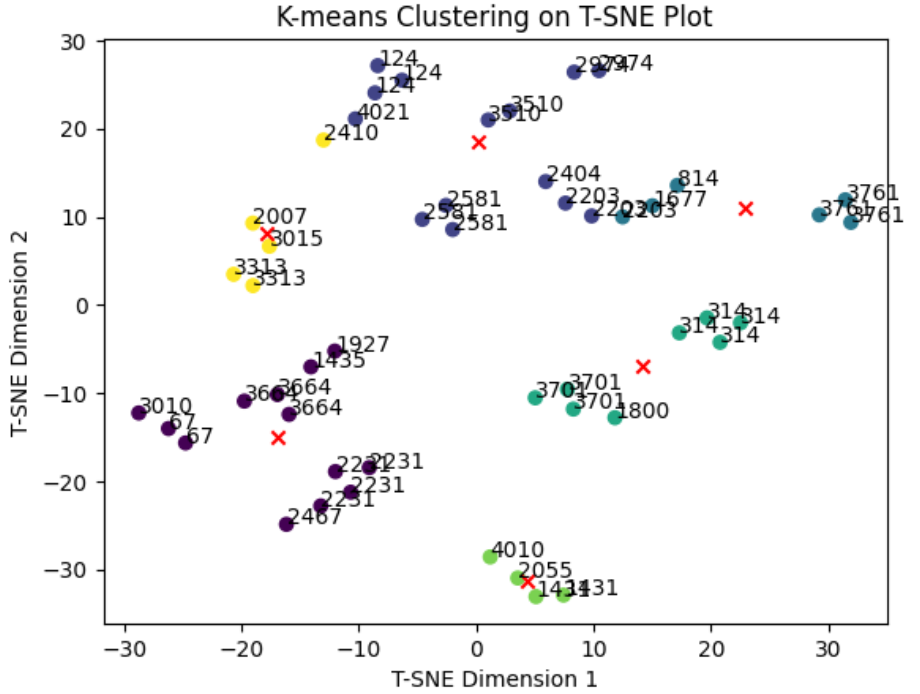}}
  \caption{K-means Clustering on t-SNE plot for different fine-tuned model's image outputs in same latent space obtained by the query ”Give me polo neck t-shirt for men”} 
  \label{16}
\end{figure}

Based on Algorithm \ref{6}, Fig.~\ref{16} represents a step of performing K-means Clustering on the t-SNE plot for different fine-tuned model's image outputs in the same latent space obtained by the query ”Give me polo neck t-shirt for men”. It is done to obtain the head cluster which has the head image(the most frequent image in different fine-tuned model's image output results). The K-means clustering algorithm is employed to effectively group data points that exhibit similarity, based on a predefined number of clusters. Applying K-means on the t-SNE plot helps visualize model's representations and how these representations cluster together in the reduced dimensionality space. The Elbow Method is used with the Silhouette Method to find the best K value in K-Means Clustering \cite{19}. The K value between the range of 4 and 6 gave the best results.

The evaluation and ranking of images have been conducted using the following methods:

\begin{itemize}
  \item Sorting in descending order based on \textit{weighted\_score}.
  \item Sorting in descending order based on the product of frequency and \textit{weighted\_score}. 
  \item Sorting in descending order based on frequency and \textit{weighted\_score}. 

\end{itemize}
Ranking based on frequency and \textit{weighted\_score} yielded the best results in the findings.

The ensemble of multiple fine-tuned models is particularly effective in scenarios with limited data and low-quality images. Different models capture different aspects of the data, and combining them helps mitigate the limitations of any single model.
By integrating outputs from models trained for different epochs, ENCLIP approach leverages the strengths of models at various stages of training. Early epochs may generalize better due to less overfitting, while later epochs capture more refined features. This combination ensures a robust performance even with limited data
The latent space representations extracted from the models encapsulate rich, high-dimensional features of the images. These features are crucial for handling low-quality images, as they can emphasize important aspects of the data that are less affected by noise and low resolution.\par
By performing clustering in the latent space, even images of lower quality are grouped based on their most salient features rather than raw pixel values. This method is more resilient to variations in image quality.
Assigning weights and frequencies to images based on their occurrences in multiple model outputs helps in identifying consistently important images. This is particularly useful when dealing with limited and low-quality data, as it highlights images that are recognized as important across different models.
The weighted score formula gives higher importance to images appearing in models trained for more epochs, reflecting improved learning. This prioritization helps in selecting images that are likely to be of higher relevance and quality despite the overall data limitations.

In FashionCLIP \cite{9}, for final training around 700k high-resolution product images were used which was provided by Farfetch. In the ENCLIP approach, approximately 35k low-resolution product images were used for training, which is significantly less—20 times fewer—than the Farfetch dataset used in FashionCLIP.

\section{Conclusion}
The introduction of multimodal search has brought about a revolutionary change in the fashion industry by providing users with a seamless and intuitive way to explore and discover fashion items. By integrating text and image information, users can now search for products based on their preferences, style, or specific attributes. This advancement has greatly enhanced the overall user experience and opened up new possibilities for fashion discovery. This paper presents an innovative approach called ENCLIP, specifically tailored to enhance the fine-tuning performance of the Contrastive Language-Image Pretraining (CLIP) model in the domain of fashion intelligence. The ENCLIP approach addresses the challenges posed by limited data availability and low-quality images, which are prevalent in the fashion industry. In ENCLIP approach, the process involves training and ensembling multiple instances of the CLIP model. Additionally, a clustering technique is employed to group similar images together to get the top N images based on the user's query for searching the top relevant N images. \par
The efficacy of the ENCLIP approach is demonstrated through experimental results, comparing it with recent studies. By leveraging the clustering techniques and training ensembles of CLIP models, the method successfully overcomes the limitations of data scarcity and low-quality images. However, it is important to note some limitations of the current approach. The ENCLIP approach may not be well-suited for fine-grained querying in the case of text-to-image search, where users require highly specific search results. The model's generalization capabilities may be limited in such cases, and further research is needed to improve its performance in fine-grained fashion queries. Performance in the case of fine-grained queries could be improved by adding more textual descriptions to enhance textual understanding in terms of data quality or by exploring more advanced multimodal fusion techniques\cite{20}. The advancements made in this paper pave the way for further advancements in fashion intelligence and contribute to the ongoing evolution of multimodal search technology.

\subsubsection{Acknowledgements} Thanks are extended to the anonymous reviewers and participants for their time and feedback. Sincere gratitude is also expressed to Lalit Bhise, founder and CEO of Bizom (Mobisy Technologies Private Limited), for his enthusiasm and invaluable support for this project.

%
%
%
\bibliographystyle{splncs04}

\end{document}